\documentclass[runningheads]{llncs}
\usepackage[T1]{fontenc}
\usepackage{graphicx}
\usepackage{amsmath,amsfonts}
\usepackage{hyperref}
\begin{document}
\title{DepthMOT: Depth Cues Lead to a Strong Multi-Object Tracker}
%
%
\author{Jiapeng Wu\inst{1} \and 
Yichen Liu\inst{2}}
\authorrunning{J. Wu et al.}
%
\institute{Beijing Institute of Technology \and 
University of Chinese Academy of Sciences}
\maketitle              
\begin{abstract}
 
Accurately distinguishing each object is a fundamental goal of Multi-object tracking (MOT) algorithms. However, achieving this goal still remains challenging, primarily due to: (i) For crowded scenes with occluded objects, the high overlap of object bounding boxes leads to confusion among closely located objects. Nevertheless, humans naturally perceive the depth of elements in a scene when observing 2D videos. Inspired by this, even though the bounding boxes of objects are close on the camera plane, we can differentiate them in the depth dimension, thereby establishing a 3D perception of the objects. (ii) For videos with rapidly irregular camera motion, abrupt changes in object positions can result in ID switches. However, if the camera pose are known, we can compensate for the errors in linear motion models. In this paper, we propose \textit{DepthMOT}, which achieves: (i) detecting and estimating scene depth map \textit{end-to-end}, (ii) compensating the irregular camera motion by camera pose estimation. Extensive experiments demonstrate the superior performance of DepthMOT in VisDrone-MOT and UAVDT datasets. The code will be available at \url{https://github.com/JackWoo0831/DepthMOT}.

\end{abstract}
\section{Introduction}

Multi-object tracking (MOT) is a fundamental task in computer vision, which is widely applied in autonomous driving, smart traffic monitoring, surveillance, etc. It aims at labeling interest objects in a video and forming trajectories. Many methods follow the tracking-by-detection (TBD) paradigm \cite{sort,deepsort,botsort,ocsort,fairmot}, which divide MOT into two steps: detection and association. In detection step, 2D bounding boxes of objects are obtained by a detector. Then in association step, motion and/or appearance features are utilized to match detections with trajectories by minimum cost rule. 

However, maintaining precise distinction among mutually occluded objects in crowded scenes still remains a challenge for MOT. This is attributed to the substantial overlap of object bounding boxes in the camera image plane, resulting in similar motion features of objects. Additionally, considering noise and disturbances by detector or Kalman Filter \cite{kalman}, confusion and ID switches may occurred among mutually occluded objects, consequently reducing the accuracy of association. It is noteworthy that relying solely on appearance features is not an optimal choice, as appearance features are prone to distortion under occlusion conditions.

\begin{figure}[!t]
\includegraphics[width=\textwidth]{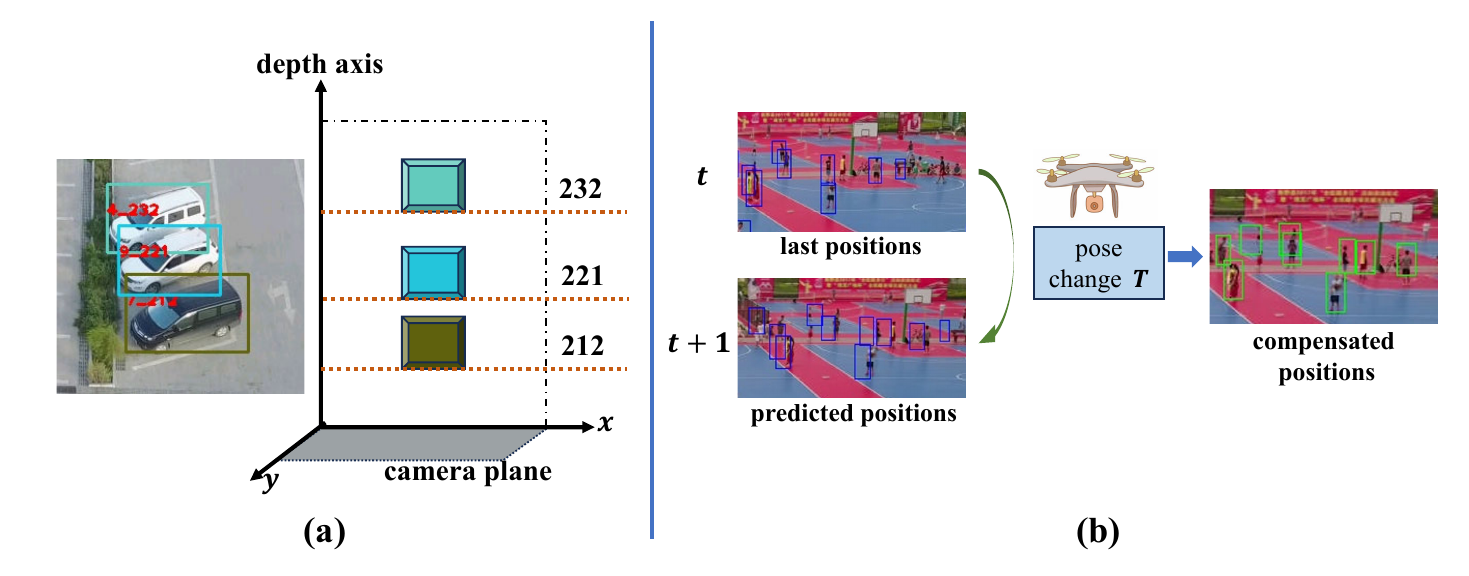}
\caption{Motivation of our work. (a) When objects occlude each other, we can distinguish them by the depth information. (b) Since depth estimation requires information about the camera pose, concurrently with depth estimation, we can also correct errors in the linear motion model (such as Kalman Filter) under irregular camera motion by changes in camera pose.} \label{fig1}
\end{figure}

By rethinking how humans track objects when watching a 2D video, it can be observed that human observations are less prone to disturbance when objects overlap. This is primarily because humans can perceive the object depth to the camera, or in other words, the 3D information. Additionally, humans can remember the "before and after" order of objects. 
Unfortunately, most of existing methods \cite{bytetrack,motr,fairmot,cstrack} represent objects solely with 2D bounding boxes, thereby losing the stereoscopic information of objects and diminishing their discriminative capability, as illustrated in Fig. \ref{fig1} (a).

Moreover, irregular camera motion raises another challenge for MOT. The nonlinear motion of the camera (acceleration/deceleration, rotation, etc.) induces nonlinear motion of objects in the frame, rendering the errors of linear models such as the Kalman Filter \cite{kalman}. To address this issue, BoT-SORT \cite{botsort} and subsequent works \cite{strongsort,sparsetrack} employ image registration methods to compute affine transformations between adjacent frames, thereby correcting the predictions of the Kalman Filter. However, this approach tends to be slow and may perform inferior accuracy in images with complex textures.

Compared to the complex computation of affine matrices, if it is possible to estimate the changes in the camera pose in the world coordinate system, the correspondence between pixels in the previous and current frames can be obtained \cite{monodepth}. In other words, if an object is predicted to "should" appear at a location by motion model prediction, but due to the camera's motion, the actual position deviates significantly. In this case, knowing "how" the camera is moving allows us to correct the estimated position. Fortunately, the estimation of the camera's pose is also essential for scene depth estimation, making addressing these two challenges mutually beneficial, as illustrated in Fig. \ref{fig1} (b).

To achieve the two goals above, we propose \textit{DepthMOT}, which detects the objects and estimates the depth map of the scene with an end-to-end manner. Also, by predicting the camera pose, error of linear motion model is compensated under irregular camera motion, especially in drone-captured videos. 

Specifically, We adopt FairMOT \cite{fairmot} as our baseline. After obtaining multi-scale image features through the Encoder (DLA-34), these features are input into a \textit{depth branch} composed of a U-Net \cite{unet}-like structure to estimate the scene depth. The depth of the objects are then calculated by the average depth of pixels covered by the bounding box. Since there are no ground truth depth values in MOT datasets \cite{visdrone,uavdt}, we resort to self-supervised monocular depth estimation methods \cite{monodepth}, which concurrently require predicting the camera pose. To address this, we construct an additional \textit{pose branch}. The pose branch takes two adjacent frames as inputs and estimates the camera's 6-DoF. During the inference phase, we decompose the association sequence based on the depth of objects and utilize the estimated 6-DoF to compensate for the errors by Kalman Filter effectively.

In summary, our contributions are as follows:

- We propose DepthMOT, which breaks the gap between Multi-object tracking and self-supervised monocular depth estimation.

- Achieving 3D perception of objects by estimating their depth, thereby distinguishing mutually occluded objects in dense scenes.

- Correcting errors in the linear motion model under irregular camera motion by estimating changes in the camera's pose.

\section{Related Work}

\subsection{Multi-Object Tracking}

Mainstream MOT methods can be categorized into two paradigms, namely tracking-by-detection (TBD) and joint detection and tracking (JDT). In TBD paradigm, an off-the-shelf detector (like \cite{yolox}) is firstly utilized to generate bounding boxes, and identity of objects is determined by an association strategy \cite{sort,deepsort,bytetrack,botsort,ocsort}. However, the discrete separation of detection and tracking processes can sometimes result in redundant computations, especially in cases where appearance features are required. In JDT paradigm, the model simultaneously outputs detection results and clues related to tracking. For example, JDE \cite{jde}, FairMOT \cite{fairmot} and CSTrack \cite{cstrack} generate bounding boxes and feature embeddings in a singel forward. TraDes\cite{trades} estimate the position offset related to objects concurrently with detection results. Besides, Transformer \cite{transformer}-like methods \cite{trackformer,transcenter,gtr,memot} directly process current potential detections and historical trajectories into query embeddings, thereby outputting trajectories directly without explicit association processes.

\subsection{Depth Estimation}

Due to the relatively high cost of sensors and expense of annotation, pure visual self-supervised depth estimation has garnered increasing attention. In comparison to stereo estimation, although monocular estimation is more challenging, it exhibits better transferability. Typically, it employs an encoder (such as the UNet \cite{unet} architecture) for dense depth estimation, while requiring another network (often referred to as PoseNet) to estimate the camera pose. By computing the depth and pose changes between consecutive frames in a video, constraints on depth estimation can be imposed through image structural losses.

Monodepth2 \cite{monodepth} use the re-projection loss and auto-masking stragegy to solve the issue of occlusion and static camera scenes, which is widely used as a baseline. HRDepth \cite{hrdepth} redesigned the encoder to predict high-resolution maps with a low complexity. SC-Depth v3 \cite{scdepthv3} use external pretrained monocular depth estimation model for generating single-image depth prior in dynamic scenes. Additionally, some methods like \cite{Semandepth} introduces the semantic segmentation network to guide depth estimation. 

\subsection{Depth cues in MOT}
 
Currently, in 2D MOT community, the utilization of depth information has not yet sparked widespread discussion. To our best knowledge, there are primarily two works that integrate MOT with depth estimation. QuoVadis \cite{quovadis} employs depth estimation and semantic segmentation to construct a point cloud of the ground in the scene. By combining the 2D detection results from a detector, it estimates the position of objects in the Bird's Eye View (BEV) perspective for trajectory prediction. However, it comes with a significant computational complexity. SparseTrack \cite{sparsetrack} does not use network-estimated depth but instead leverages the principle of "closer objects appear larger". It estimates the depth of objects as the distance from the bottom of the bounding box to the bottom of the image. However, this method is sometimes inaccurate, as it does not truly integrate with scene information and may result in "parallax illusions."

\section{Methodology}

\subsection{Preliminaries}

\begin{figure}[!t]
\includegraphics[width=\textwidth]{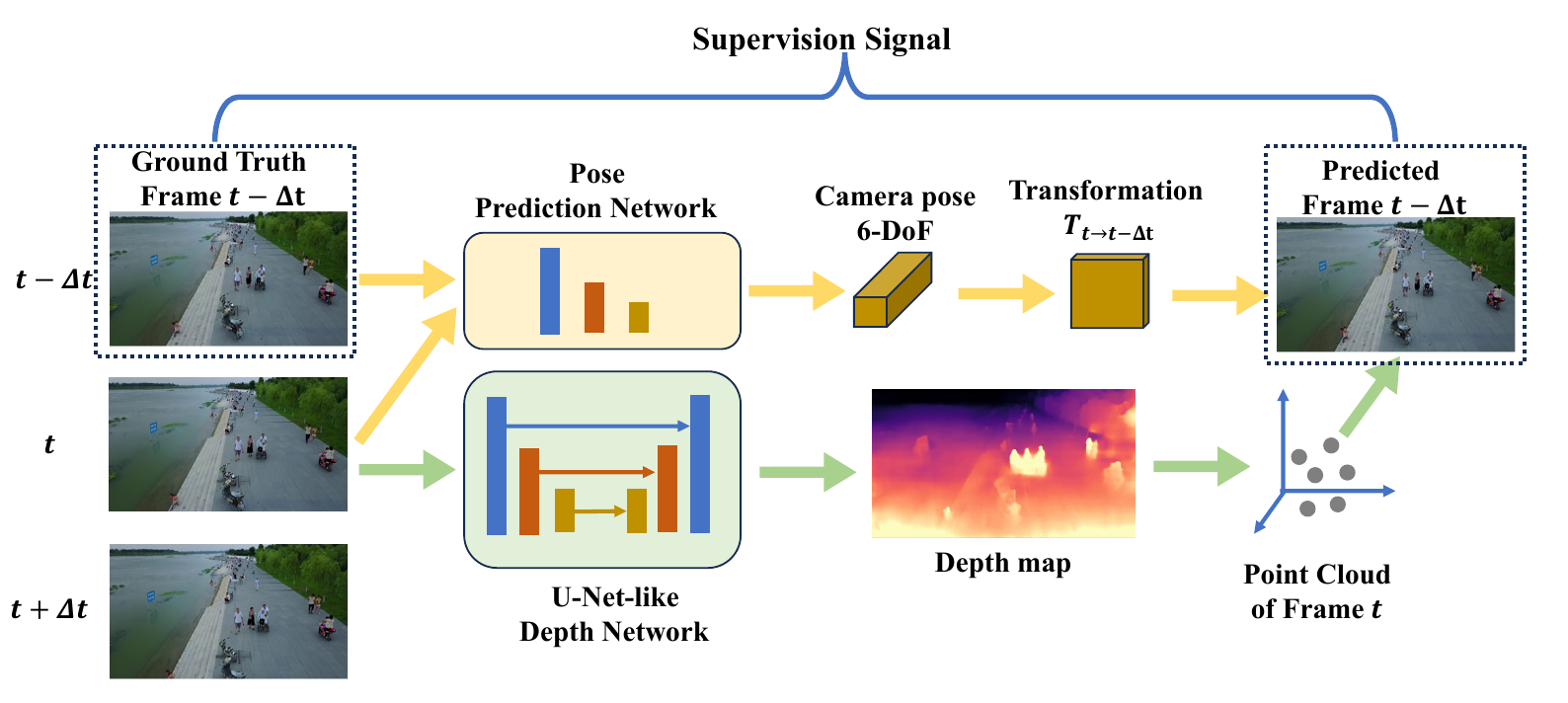}
\caption{Flowchart of training process of self-supervised monocular depth estimation.} \label{fig2}
\end{figure}

For a better understanding of our work, we firstly introduce the basic components and steps of self-supervised monocular depth estimation, as shown in Fig \ref{fig2}. 

Generally, the model takes consecutive frames in a video as input, such as the frames $t - \Delta t, t, t + \Delta t$ in Fig. \ref{fig2}. Subsequently, a U-Net \cite{unet}-like network performs a dense prediction task, estimating the depth map of the intermediate frame (frame $t$).

At this point, we do not know whether the predicted depth map is accurate. To address this, we concatenate the images of two consecutive frames (e.g., frames $t - \Delta t, t$) and feed them into a pose estimation network. This network outputs the 6-DoF camera pose change between these two frames. Based on 6-DoF, the transformation between the two frames $\mathrm{T}_{t \to t - \Delta t} = \{R, \tau \}$ is calculated, where $R$ is the rotation matrix and $t$ is the translation vector. 

Then, for pixel $p = [x, y, d]^T $ in frame $t$ (where $d$ is the depth predicted by depth network), we can infer its new position $p'$ in frame $t - \Delta t$ by Eq. \ref{eq: 1}:

\begin{equation} \label{eq: 1}
    p' = K R K^{-1} p + K^{-1} \tau 
\end{equation}

\noindent where $K$ is the camera intrinsic matrix. Therefore, we can reconstruct $t - \Delta t$ according to $t$, and then compare it with the ground truth image to use as the supervisory signal for training. 

\subsection{DepthMOT}

Our goal is to end-to-end learn tracking and depth information within the same network. To achieve this goal, we choose FairMOT \cite{fairmot} as baseline and integrate the components for depth estimation elegantly. The overall architecture of DepthMOT is shown in Fig. \ref{fig3}:

\begin{figure}[!t]
\includegraphics[width=\textwidth]{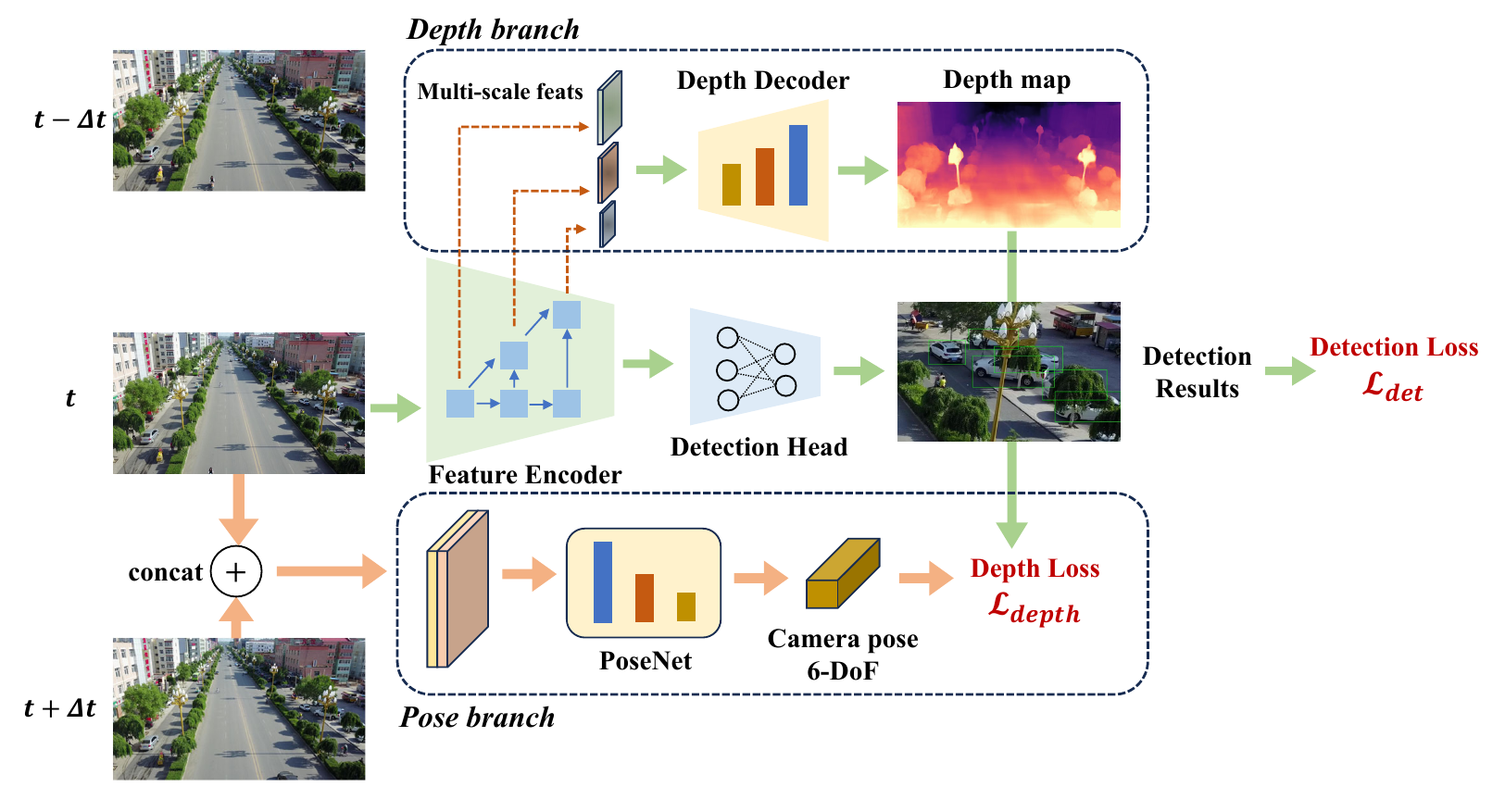}
\caption{Diagram of our proposed DepthMOT} \label{fig3}
\end{figure}

Similar to FairMOT \cite{fairmot}, we utilize DLA-34 \cite{centertrack} as the backbone to extract image features for the current frame $I_t$, represented as 
$f_t \in \mathbf{R}^{c \times h \times w}$, where $h = H / 4, w = W / 4$, $H, W$ being the input image size.

The detection branch comprises three heads: heatmap, bounding box size, and center offsets. Each head consists of 2D convolutional layers. The heatmap head is responsible for predicting the center of objects, the bounding box size head predicts the height and width of the bounding box, and the center offsets head fine-tunes the position of the center, resisting the errors introduced by the downsampling. This part is consistent with FairMOT \cite{fairmot}.

The core components of DepthMOT are Depth branch and Pose branch, described as follows:

\subsubsection{Depth Branch} The DLA-34 backbone network includes multiple operations of multi-scale feature fusion \cite{centertrack,fairmot}, allowing it to obtain more detailed features compared to ResNet-50 \cite{resnet}, which is commonly used in most depth estimation works \cite{monodepth,hrdepth}. Specifically, within DLA-34, there are four Tree structures, and we obtain features at the head, tail, and middle, resulting in a total of five scales: $\mathbf{F} = \{f_i\}_{i=1}^5, f_i \in \mathrm{R}^{2^i c_0 \times H / 2^i \times W / 2^i}$, where $c_0 = 16$.

Subsequently, the multi-scale features are fed into Depth Decoder to predict multi-resolution depth maps. Specifically, for $s$-th scale, the depth map  is predicted by:

\begin{equation} \label{eq: 2}
    D_s = \mathrm{Sigmoid}(\mathrm{Conv}(f_s, \mathrm{Upsample}(f_{s + 1}))) \in \mathrm{R}^{1 \times H / 2^s \times W / 2^s}
\end{equation}

\noindent where $\mathrm{Conv}$ indicates a set of 2D convolution layers. It worth noticing that the depth map $D_s$ in \ref{eq: 2} is actually disparity map, which means the distance between the piont and camera plane. As for the "real" depth, given minimum depth constraint $\tilde{d}_{min}$ and maximum constraint $\tilde{d}_{max}$, it could be calculated by:

\begin{equation} \label{eq: disp}
    \tilde{d} = \frac{1}{1 / \tilde{d}_{max} + (1 / \tilde{d}_{min} - 1 / \tilde{d}_{max}) d}
\end{equation}

For convenience, we use $\tilde{d}, d$ to refer the "real" depth and disparity respectively in the remaining part  of the paper.

\subsubsection{Pose Branch} We construct Pose Branch like most of the depth estimation work \cite{monodepth,hrdepth,scdepthv3,radepth}. We concatenate temporally adjacent frames along the channel dimension and then feed them into ResNet-18 \cite{resnet} to obtain feature maps. Subsequently, through several layers of 2D convolutional layers, we predict the 6-DoF camera pose (rotation angles and translation) $p = [\theta_x, \theta_y, \theta_z, t_x, t_y, t_z]^T$.

\subsection{Training}

As shown in Fig. \ref{fig3}, the traning loss can be divided into two parts: Detection Loss $\mathcal{L}_{det}$ similar to FairMOT \cite{fairmot} and Depth Loss $\mathcal{L}_{depth}$ regarding to the depth estimation subtask.

\subsubsection{Detection Loss}

Keeping same as the FairMOT \cite{fairmot}, the detection loss $\mathcal{L}_{det}$ is composed of heatmap loss $\mathcal{L}_{heat}$ and box size loss $\mathcal{L}_{box}$.

The output heatmap represents the probability of existence of objects. Therefore, a ground truth heatmap defined by Gaussian distribution can be obtained by:

\begin{equation} \label{eq: 3}
    M[x, y] = \sum_{n=1}^N \exp \{ - \frac{(x - x_{c, n})^2 + (y - y_{c, n})^2}{2 \sigma^2} \}
\end{equation}

\noindent where $N$ represents the number of objects and $(x_{c, n}, x_{c, n})$ represents the center position of $n$-th GT bounding box corresponding to the heatmap scale. 

Therefore, the heatmap loss is defined as focal loss \cite{focalloss}:

\begin{align} \label{eq: 4}
    \mathcal{L}_{heat} = \begin{cases}
        - \frac{1}{N} \sum_x \sum_y (1 - \tilde{M}[x, y]) ^ \alpha \log (\tilde{M}[x, y]) , & M[x, y] = 1 \\
        - \frac{1}{N} \sum_x \sum_y (1 - \tilde{M}[x, y]) ^ \beta \tilde{M}[x, y]^\alpha \log (\tilde{M}[x, y]), & otherwise
    \end{cases}
\end{align}

where $\alpha$ and $\beta$ are hyper parameters, $\tilde{M}$ is the estimated heatmap.

The box size loss $\mathcal{L}_{box}$ is defined by $L_1$-norm difference of predicted and ground truth center positions and shapes:

\begin{equation} \label{eq: 5}
    \mathcal{L}_{box} = \sum_{n = 1}^N ( \Vert [x_{c, n}, y_{c, n}] - [\tilde{x}_{c, n}, \tilde{y}_{c, n}] \Vert + 0.1 \Vert[w_n, h_n] - [\tilde{w}_n, \tilde{h}_n] \Vert)
\end{equation}

\noindent where tilde means the estimated value. Finally, 

\begin{equation} \label{eq: 6}
    \mathcal{L}_{det} = \mathcal{L}_{heat} + \mathcal{L}_{box}
\end{equation}

\subsubsection{Depth Loss}

As mentioned in Sec. 3.1, the self-supervised depth estimation is mostly supervised by reconstruction or reprojection errors. Specifically, at timestamp $t$ and scale index $s$, firstly we upsample the estimated depth map $D_s$ by Depth Branch to the original scale, i.e., $\tilde D_s \leftarrow \operatorname{Upsample} (D_s) \in \mathrm{R}^{1 \times H \times W}$. Subsequently, since the camera pose change between nearby frames ($t$ and $t'$, $t' = t \pm \Delta t$) is also predicted by Pose Branch, we can obtain the reprojected $t'$-th frame $\tilde{I}_{t'}$ by Eq. (\ref{eq: 1}). 

Following \cite{monodepth},structural similarity \cite{ssim} and $L_1$-norm are utilized to measure the difference of original $t'$-th frame $I_{t'}$ reprojected one $\tilde{I}_{t'}$ according to Eq. (\ref{eq: 7}):

\begin{equation} \label{eq: 7}
    pe(I_{t'}, \tilde{I}_{t'}) = \frac{\alpha}{2} (1 - \operatorname{SSIM}(I_{t'}, \tilde{I}_{t'})) + (1 - \alpha) \Vert I_{t'} - \tilde{I}_{t'} \Vert
\end{equation}

The reprojection loss can be defined by pixel-wise minimum of $pe(I_{t+\Delta t}, \tilde{I}_{t+\Delta t})$ and $pe(I_{t-\Delta t}, \tilde{I}_{t-\Delta t})$ to handle pixel occlusion, that is:

\begin{equation} \label{eq: 8}
    \mathcal{L}_p = \min_{t'} pe(I_{t'}, \tilde{I}_{t'}), \quad t' \in \{t+\Delta t, t-\Delta t\}
\end{equation}

Also, aiming to smooth the generated depth map, we avoid sharp textures by minimizing the first-order gradients in the x and y directions, as shown in Eq. (\ref{eq: 9})

\begin{equation} \label{eq: 9}
    \mathcal{L}_s = |\partial_x D_s| e^{-|\partial_x I_t|} + |\partial_y D_s| e^{-|\partial_y I_t|}
\end{equation}

Finally, depth loss is the sum of reprojection loss and smooth loss in every scale: 

\begin{equation} \label{eq: 10}
    \mathcal{L}_{depth} = \sum_{i = 1}^S \mathcal{L}_p^i + \lambda \mathcal{L}_s^i
\end{equation}

\noindent where $\lambda = 0.001$, $S = 5$ in our design.

\subsubsection{Overall Loss}

To comprehensively balance the two subtasks: detection and depth estimation, following \cite{fairmot}, we leverage the Uncertainty loss \cite{uncertaintyloss} to automatically adjust the weights of two loss functions, as illustrated in Eq. (\ref{eq: 11}):

\begin{equation} \label{eq: 11}
    \mathcal{L} = 0.5( e^{-w_1}\mathcal{L}_{det} + e^{-w_2}\gamma \mathcal{L}_{depth} + w_1 + w_2)
\end{equation}

\noindent where $w1, w2$ are the learned weights and $\gamma = 50$ in default to align the value scale between depth loss and detection loss.

\subsection{Inference}

Under the circumstance of dense distribution of objects or occlusion, the bounding boxes are closed to each other and objects are prone to be confused, resulting in ID Switches. To alleviate this issue, depth information of scene is estimated to form a 3D perception of objects. Specifically, donating the depth map of current frame $t$ as $D \in \mathrm{R}^{1 \times H \times W}$, where $H, W$ is the original size of frame image. Note that although the model predicates multi-scale of depth maps, only the maximum scale, i.e. the original image scale, is adopted.

For each object $o^i = [x_0, y_0, x_1, y_1]$, which obeys the top-left-bottom-right format, the depth is defined by:

\begin{equation} \label{eq: 12}
    d_i = \frac{1}{x_1 - x_0} \sum_{x = x_0}^{x_1} D[x, y_1]
\end{equation}

Eq. (\ref{eq: 12}) means the depth of an object is determined by the average of depth value in the bottom edge of bounding box. Since the camera is typically positioned in an overhead view, the bottom edge of the bounding box often aligns with the ground, providing a good representation of the object's depth.

After obtaining depth values of each object, following SparseTrack \cite{sparsetrack}, we adopt a cascade manner to associate detections with trajectories. To be specific, the depth values of all detections and trajectory objects are divided into several evenly spaced intervals, and then the trajectories and detections with the corresponding depth level are matched by IoU metric. Each round of unsuccessful matches for detections or trajectories is carried over to the next round for matching. The benefit of cascading matching based on depth values is that it allows separating the matching of objects with high bounding box overlap, thereby enhancing discrimination.

As for the compensation of errors caused by irregular camera motion, as mentioned before, we could predict the position that objects "should" appear by camera pose changes. For example, the position of object $o^i$ in frame $t$ is $[x_0, y_0, x_1, y_1]$ predicted by Kalman Filter \cite{kalman}. It worth noticing that since the bottom line of bounding boxes are always on the ground, we only compensate the bottom-left and bottom-right points ($[x_0, y_1]$ and $[x_1, y_1]$) of bounding boxes, and maintain the height of box unchanged. In specific, from timestamp $t - 1$ to $t$, 6-DoF of camera motion can be predicted by Pose Branch. Thus, the coordinate transformation $\mathrm{T}_{t - 1 \to t} = \{R, \tau \}$ is further obtained, and the new position of $p = [x_i, y_j, \tilde{d}]$ is as Eq. (\ref{eq: 13}):

\begin{equation}
    [x_i, y_j, \tilde{d}]^T = KRK^{-1} [x_i, y_j, \tilde{d}]^T + K^{-1} \tau , \quad i = 0, 1; \quad   j = 0
\end{equation} \label{eq: 13}

\noindent where $\tilde{d}$ is the corresponding real depth.

\section{Experiments}

\subsection{Datasets and Metrics}

Based on the hypothesis of self-supervised monocular depth estimation, only the datasets that possess relatively obvious camera motion are adopted. Therefore, we choose two popular UAV multi-object tracking datasets, i.e., VisDrone \cite{visdrone} and UAVDT \cite{uavdt}, to conduct experiments. 

VisDrone \cite{visdrone} contains 56 videos for training, 7 videos for validation and 17 sequences for testing. Additionally, it includes various scenarios: car, bus, truck, pedestrian, and van. UAVDT \cite{uavdt} contains totally 50 sequences, and only cars are considered to train and test. Following \cite{stntrack}, we randomly choose 40 sequences for training and the rest for testing.

To comprehensively evaluate the performance of our proposed method, both Higher-Order Tracking Accuracy (HOTA) \cite{hota} and classical CLEAR metrics (including MOTA, IDF1, MT, ML, IDs, etc.) \cite{clear} are utilized. Typically, we mainly consider HOTA, MOTA and IDF1. HOTA reflects the overall performance of identity stability and detection, while MOTA mainly represents the number of incorrect detections (False positives and False negatives) and ID Switches. IDF1 means the association performance. 

\subsection{Implementation Details}

As illustrated before, we take FairMOT \cite{fairmot} as our baseline. As for Depth Branch, we mainly follow the structure of monodepth2 \cite{monodepth}. The last five feature maps from DLA-34 backbone are extracted to feed in Depth Decoder. Compared to origin ResNet-18 \cite{resnet} used in monodepth2 \cite{monodepth}, multi-scale features extracted in DLA-34 possess a stronger representational capabilities. We train 10 epochs for VisDrone \cite{visdrone} dataset and 20 epochs for UAVDT \cite{uavdt} dataset with a single Tesla A100 GPU. The learning rate is set to $10^{-4}$ as same in FairMOT \cite{fairmot}, with a batch size of 4. The optimizer is chosen as Adam and the size of input image is set to $1088 \times 608$. 

\subsection{Main Results}

In this part, we compare the performance of DepthMOT with popular state-of-the-art methods which folow tracking-by-detection (TBD) paradigm \cite{sort,deepsort,bytetrack,ocsort,botsort,uavmot} or joint detection and tracking (JDT) paradigm \cite{fairmot,trackformer}. Note that YOLOX-m \cite{yolox} detector is chosen as the detector for TBD methods, and is trained for 35 epochs for VisDrone \cite{visdrone} and 50 epochs for UAVDT \cite{uavdt}. As for the rest JDT methods, we re-trained them in the two datasets following the configurations corresponding to the best performance.

\subsubsection{VisDrone-MOT dataset} 

The result is shown in Table \ref{result_visdrone}. It shows that our proposed DepthMOT get the best performance in HOTA, IDs and MT metrics, meaning that DepthMOT possesses the capacity of stable tracking. Besides, the results on MOTA and IDF1 metric is comparable to other effective SOTA methods like \cite{bytetrack,uavmot}. 

\begin{table}
\caption{Comparison with SOTA MOT methods on VisDrone-MOT \cite{visdrone} dataset. The best results are indicated in bold.} \label{result_visdrone}
\begin{tabular}{ c | c c c c c c c c }
\hline
Method                 & HOTA$\uparrow$ & MOTA$\uparrow$ & IDF1$\uparrow$ & FN$\downarrow$ & FP$\downarrow$ & IDs$\downarrow$ & MT$\uparrow$ & ML$\downarrow$ \\
\hline
SORT \cite{sort}            & 35.08    & 33.148 & 42.844   & 112481 & 20392  & 3525   & 329 & 523 \\
DeepSORT \cite{deepsort}    & 36.921   & 34.397 & 46.714   & 110989 & 21077  & 1784   & 386 & 517 \\
ByteTrack \cite{bytetrack}  & 40.661   & \textbf{39.541} & 50.398   & 105518 & 16257  & 1581   & 507 & 538 \\
ByteTrack+ReID \cite{bytetrack} & 41.422   & 40.88  & 51.264   & \textbf{103245} & 15254  & 1512   & 510 & 525 \\
BoT-SORT \cite{botsort}     & 42.42    & 41.652 & \textbf{56.843}   & 103505 & 14114  & 1430   & 543 & 537  \\
UAVMOT \cite{uavmot}        & 38.133   & 38.685 & 45.15    & 108134 & 13610  & 3357   & 463 & 557  \\
FairMOT \cite{fairmot}      & 31.102   & 12.806 & 37.745   & 114834 & 59997  & 3072   & 397 & 581 \\
TrackFormer \cite{trackformer}  & 35.344   & 25     & 51.0     & 141526 & 25856  & 1534   & 515 & 946 \\
\textit{DepthMOT}           & \textbf{42.448}   & 37.041 & 54.023   & 104054 & 41001  & \textbf{1248}   & \textbf{626} & 467  \\
\hline
\end{tabular}
\end{table}

\subsubsection{UAVDT dataset}

We also compare our method with other SOTA methods on UAVDT \cite{uavdt} dataset, which is shown in Table \ref{result_uavdt}. However, out DepthMOT get inferior results in this round, especially in MOTA, IDF1 and FN metrics. The reason may be the difference among the detectors. Nevertheless, our method still get the best score in HOTA and FP, which represents the effectiveness.

\begin{table}
\caption{Comparison with SOTA MOT methods on UAVDT \cite{uavdt} dataset. The best results are indicated in bold.} \label{result_uavdt}
\begin{tabular}{ c | c c c c c c c c }
\hline
Method                 & HOTA$\uparrow$ & MOTA$\uparrow$ & IDF1$\uparrow$ & FN$\downarrow$ & FP$\downarrow$ & IDs$\downarrow$ & MT$\uparrow$ & ML$\downarrow$ \\
\hline
SORT \cite{sort}                & 60.4     & 66.414 & 77.119   & 19034  & 6505   & 160  & 201 & \textbf{36} \\
DeepSORT \cite{deepsort}        & 61.97    & 68.498 & 78.615   & 20035  & 4008   & \textbf{61}   & 179 & 43 \\
ByteTrack \cite{bytetrack}      & 62.179   & 68.754 & 78.76    & 20010  & 3796   & 102  & 182 & 42 \\
ByteTrack+ReID \cite{bytetrack} & 61.621   & 68.365 & 77.205   & 18067  & 6021   & 118  & \textbf{196} & 44 \\
BoT-SORT \cite{botsort}         & 61.369   & 67.741 & 78.508   & 20296  & 4323   & 64   & 181 & 43 \\
UAVMOT \cite{uavmot}            & 61.22    & 67.901 & 78.084   & 19371  & 5121   & 69   & 190 & 42 \\
BIoU\_Tracker \cite{biou}       & 62.937   & \textbf{70.323} & \textbf{79.622}   & \textbf{17405}  & 5224   & 79   & 200 & 40 \\
MOTDT \cite{motdt}              & 61.82    & 66.525 & 77.77    & 17760  & 5825   & 76   & 175 & 37 \\
FairMOT \cite{fairmot}          & 49.126   & 51.189 & 66.471   & 33102  & 4136   & 110  & 107 & 87 \\
TrackFormer \cite{trackformer}  & 43.165   & 37.924 & 53.343   & 45197  & 5585   & 680  & 67 & 75  \\
\textit{DepthMOT}               & \textbf{66.44}    & 62.279 & 78.13    & 28951  & \textbf{3036}  & 82   & 134 & 40 \\ 
\hline
\end{tabular}
\end{table}

\subsubsection{Visualization results}

In this part, we visualize some challenging scenes in VisDrone \cite{visdrone} and UAVDT \cite{uavdt} dataset and the predicted depth values by Depth Branch in crowded scenes to further show the effectiveness of our method.

\textbf{a) Qualitative visualization results}. 

\begin{figure}[!t]
\includegraphics[width=\textwidth]{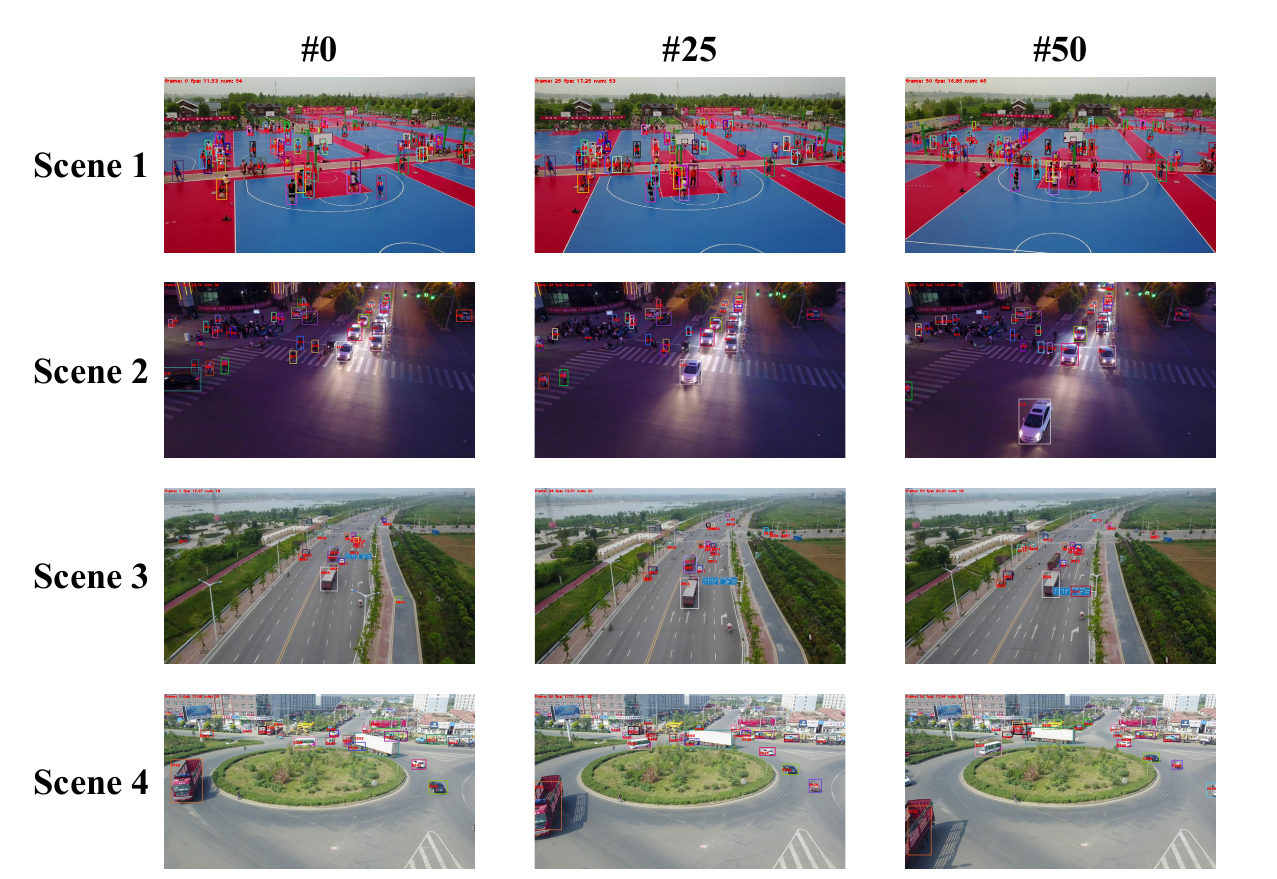}
\caption{Visualization results on challening scenes in Visdrone test set. The number on the up-left corner of bounding boxes indicates object IDs.} \label{fig4}
\end{figure}

\begin{figure}[!t]
\includegraphics[width=\textwidth]{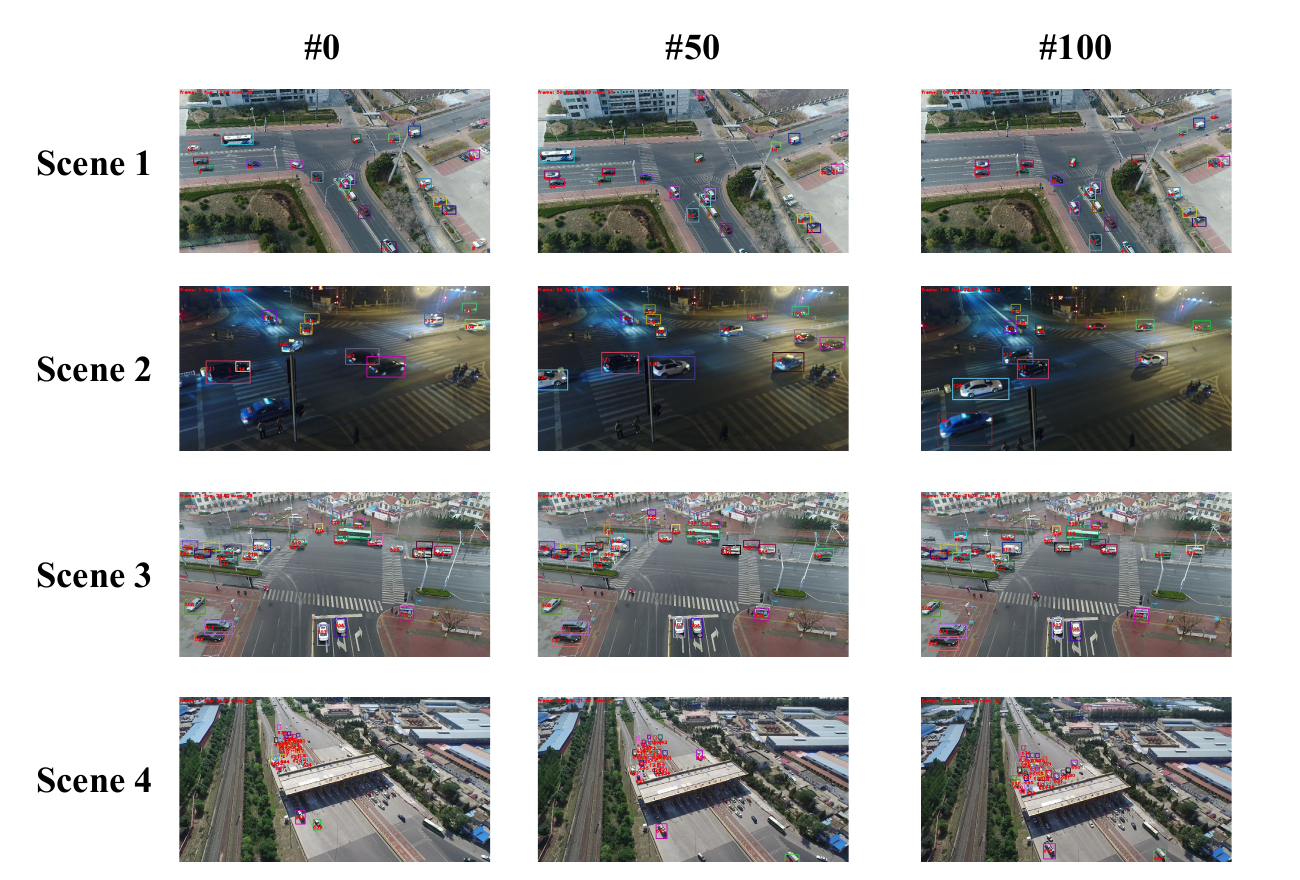}
\caption{Visualization results on challening scenes in UAVDT test set. The number on the up-left corner of bounding boxes indicates object IDs.} \label{fig5}
\end{figure}

The visualization of some challenging sequences in VisDrone \cite{visdrone} and UAVDT \cite{uavdt} datasets are shown in Fig. \ref{fig4} and Fig. \ref{fig5}, respectively.

\begin{figure}[!t]
\includegraphics[width=\textwidth]{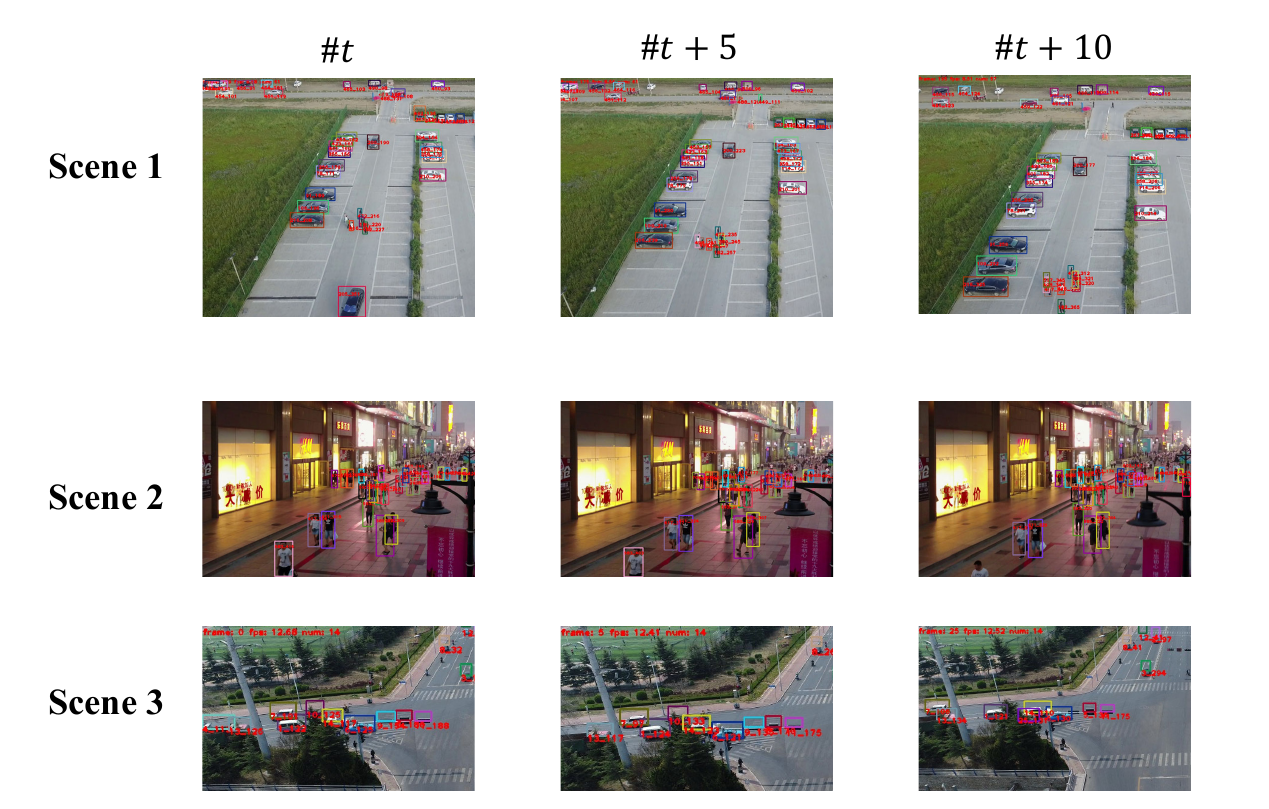}
\caption{Visualization of predicted depth values. The number on the up-left corner of bounding boxes indicates object IDs and depth value.} \label{fig6}
\end{figure}

\textbf{b) Visualization of predicted depth values}.

To validate the performance of depth estimation in crowded scenes, we visualize the predicted depth by Depth Branch (actually, the shown depth value is $10^3$ times of "disparity", as mentioned in \ref{eq: disp} ). The larger value means more closer distance between the object and camera plane. From Fig. \ref{fig6}, it can be observed that the model could predict a discriminative depth value in several adjacent objects. Also, the depth values are relatively stable, meaning that their order of magnitude can be maintained over time, so the cascade matching strategy based on depth values is reliable.

\subsection{Ablation Studies}

To fairly illustrate the effectiveness of two proposed contributions, i.e., depth cascade matching and camera motion compensation, we conduct ablation studies on VisDrone-test-dev set. Specifically, we take the origin FairMOT \cite{fairmot} as our baseline. Afterwards, we respectively 1) add low-confidence detections handling proposed in ByteTrack \cite{bytetrack}; 2) add depth cascade matching strategy proposed in SparseTrack \cite{sparsetrack}, which calculate the depth of object $o = [x_0, y_0, x_1, y_1]$ as $d = 2000 - y_1$; 3) replace 2) by our proposed depth estimation method in Eq. (\ref{eq: 12}); 4) add RANSAC algorithm proposed in BoT-SORT \cite{botsort} to compensate the irregular camera motion; 5) replace 4) by our proposed camera motion compensation (Eq. (\ref{eq: 13})). Each of the above experiments corresponds to a row in the Table \ref{result_abl}.  

\begin{table}
\caption{Ablation studies on depth cascade matching and camera motion compensation.} \label{result_abl}
\begin{tabular}{ c | c c c c  }
\hline
Operation                   & HOTA$\uparrow$ & MOTA$\uparrow$ & IDF1$\uparrow$ & IDs$\downarrow$ \\
\hline
Baseline \cite{fairmot}          & 36.112            & 33.688              & 42.573          & 5438 \\
\hline
+ByteTrack \cite{bytetrack}      & 40.089 (+3.977)   & 33.422 (-0.266)     & 49.843 (+7.27)  & 2216 (-3222) \\
\hline
+SparseTrack \cite{sparsetrack}  & 40.102 (+3.99)    & 33.249 (-0.439)     & 49.124 (+6.551) & 2721 (-2717)\\
\textit{+Our depth}  & 40.088 (+3.976) & 33.944 (+0.256) & 49.639 (+7.066) & 2689 (-2749) \\
\hline
+BoTSORT \cite{botsort}          & 40.013 (+3.901)   & 30.295 (-3.393)     & 51.216 (+8.643) & 2142 (-3296) \\
\textit{+ Our motion comp.} & \textbf{41.704 (+5.592)} & 32.622 (-1.066) & \textbf{53.635 (+11.062)} & \textbf{1381 (-4057)} \\
\hline
\end{tabular}
\end{table}

From Table \ref{result_abl}, it can be observed that our method brings a 5.5\% increase in HOTA and a 11.06\% increase in IDF1 compared to the baseline, indicating that both the proposed depth estimation module and the camera motion compensation module have contributed positively, with the latter significantly improving the stability of tracking.

\section{Conclusion and Future Work}

In this paper, we propose DepthMOT, which simultaneously detecting objects and estimating the depth map of scenes. Aiming at the frequent occlusion and crowded scenarios, although the nearby objects are easily confused in camera plane, they could be distinguished by the depth values. To achieve this goal, we estimate the depth of an object by the average depth value of the bottom line of bounding box. Also, under the rapid camera motion situation, there exists a misalignment between the predicted box and detection box. To tackle this issue, we use the camera pose change between the adjacent frames to correct the errors of Kalman Filter. 

However, there are still some setbacks of our method. Firstly, introducing the extra depth branch causes more computation cost. Besides, it also remains a challenge to accurate estimate the depth map in a self-supervised manner. Finally, how to better represent the depth of objects from depth maps is also a question worth exploring.

%
%

\bibliographystyle{splncs04}
\bibliography{ref}

\end{document}